\documentclass[cameraready]{Interspeech}

\usepackage{algorithm}
\usepackage{algorithmic}
\usepackage{mathptmx}
\usetikzlibrary{arrows.meta,positioning,automata}
\usepackage{booktabs}
\usepackage{multirow}
\usepackage{subcaption}
\usepackage{todonotes}

\title{Constrained CTC Decoding for Efficient Diacritic Restoration}

\author[orcid=0000-0001-8196-698X]{Rufael}{Marew}
\author[orcid=0000-0001-5772-9626]{Amr}{Keleg}
\author[orcid=0000-0003-1706-1777]{Hanan}{Aldarmaki}

\address{
    Mohamed Bin Zayed University of Artificial Intelligence, UAE
}

\email{\{rufael.marew,amr.keleg,hanan.aldarmaki\}@mbzuai.ac.ae}

\keywords{speech recognition, diacritic restoration}

\begin{document}

\maketitle

\begin{abstract}
In this work, we address diacritic restoration for Arabic speech transcripts. Most speech data are undiacritized, limiting the ability of modeling fine-grained phonological distinctions. The speech modality has recently been explored as a way to complement text-based diacritic restoration efforts. We propose an efficient non-autoregressive approach for speech-to-text diacritization based on Connectionist Temporal Classification (CTC). Our method incorporates hard constraints during decoding by constructing a character-level diacritization lattice from an undiacritized transcript and restricting hypotheses to valid diacritized realizations. We evaluate on Classical Arabic and Modern Standard Arabic test sets (namely, ArVoice and ClArTTS) against a more computationally-complex multi-modal diacritic restoration baseline, and show statistically significant reductions in diacritic error rates in both, demonstrating that the proposed approach offers both performance and efficiency gains.\footnote{You can access our codebase through: \url{https://github.com/rufaelfekadu/DiaCTC}}
\end{abstract}

\section{Introduction}

The Arabic script is an abjad writing system with alphabetic characters mainly representing consonants and long vowels, and additional diacritical marks representing short vowels, gemination, nunation, or the absence of a vowel. 
However, diacritics are typically dropped in text \cite{habash2010introduction}
and inferred from the surrounding context, making Arabic script under-specified for speech applications. 
In downstream pipelines, such as text-to-speech (TTS) and assistive reading interfaces, the absence of diacritics can lead to systematic pronunciation errors and ambiguity, particularly for homographs and morphologically complex forms.
Additionally, automatic speech recognition (ASR) models are typically unable to generate fully diacritized transcripts, given the lack of large-scale fully diacritized data. Training data often has \emph{mixed coverage}: some corpora are fully diacritized, such as Classical Arabic data (CA) or curated Modern Standard Arabic (MSA) resources, whereas large ASR corpora and conversational speech transcripts are typically completely undiacritized and, in less common cases, partially diacritized.
\begin{figure}[ht]
    \centering
    \footnotesize
    \addtolength{\tabcolsep}{-0.4em}
    \includegraphics[]{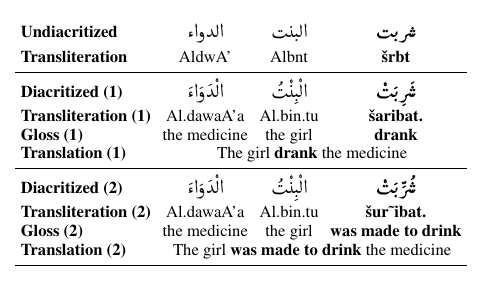}
    \caption{An example of an ambiguous undiacritized sentence with multiple possible diacritizations with English glosses and transliterations (according to the HSB scheme \cite{Habash2007}). 
    }
    \label{fig:ambiguous_example}
\end{figure}

Diacritic restoration is the task of filling in missing diacritics, which has a long history of research and development as a text-only task. In this setup, diacritic restoration is framed as a sequence-labeling task, where the input is undiacritized text, and the output is the corresponding diacritized text. One limitation of this setup is that words can still be ambiguous in context \cite{mohamed-mubarak-2025-advancing}, as exemplified in \autoref{fig:ambiguous_example}. Conversely, the presence of the corresponding speech signal could make these ambiguities recoverable from acoustic cues (e.g., vowel quality/length, gemination, case endings in careful speech, and dialectal realizations), suggesting that speech-conditioned diacritization can outperform purely text-based restoration.
Recent research demonstrated that incorporating speech data provides a complementary signal that improves diacritic restoration as the vowel information can be inferred from acoustic data  \cite{shatnawi_automatic_2024,shatnawi-etal-2024-data,ghannam2025abjad}. 
These works address \emph{speech-to-text diacritization}: given an acoustic input and an \emph{undiacritized} text reference, the goal is to predict a fully diacritized transcript that is consistent with both the speech signal and the given undiacritized reference. 
All these prior methods propose combining an acoustic or ASR model with a text-based decoder in a multi-modal learning framework. This set-up ensures that the base undiacritized alphabetic characters are unchanged, and only the diacritics are added to the transcript in a constrained manner, whereas direct ASR systems typically include character insertion, deletion, and substitution errors. While these multi-modal approaches have shown potential in Classical Arabic, their generalization to Modern Standard and Dialectal Arabic is still poor \cite{talafha-etal-2025-nadi}, and more research is needed to generalize performance.

In this work, we propose a more efficient approach for speech-based diacritic restoration, where we rely on ASR CTC decoding mechanism to meet the required linguistic constraints directly without the need for an additional text-based module and multi-modal training. Our key idea is to construct a \emph{character-level diacritization lattice} from the undiacritized transcript and restrict decoding hypotheses to paths that correspond to valid diacritized outputs. This yields a partially forced-aligned decoding graph: the model emits the reference alphabetic characters in their respective locations and only predicts diacritics at expected diacritic locations.
Unlike prior multi-modal speech and text diacritization decoders, this design preserves the simplicity and speed advantages of CTC decoding and can be integrated as a drop-in decoding constraint on top of a standard CTC-based ASR model. 
We demonstrate experimentally that this approach results in more robust and generalizable diacritic restoration performance across Classical Arabic and Modern standard Arabic compared to the baseline, while being more streamlined and efficient.

\begin{figure*}[tb]
    \centering
    
    \begin{subfigure}[t]{0.48\textwidth}
        \centering
        \includegraphics[width=0.8\linewidth]{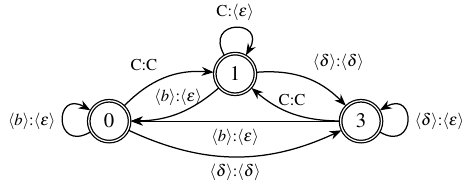}
        \caption{Default CTC topology for diacritized Arabic}
        \label{fig:ctc_default}
    \end{subfigure}
    \hfill
    \begin{subfigure}[t]{0.48\textwidth}
        \centering
        \begin{tikzpicture}[
         ->,
      >=Stealth,
      auto,
      semithick,
      node distance=7mm,
      every state/.style={minimum size=5mm, inner sep=4pt},
      lab/.style={font=\footnotesize},
    ]
    \newcommand{\eps}{\langle\epsilon\rangle}
    \newcommand{\del}{\langle\delta\rangle}
    \newcommand{\blk}{\langle b\rangle}
    
    \node[state] (q0) {0};
    \node[state, right=of q0] (q1) {1};
    \node[state, right=of q1] (q2) {2};
    \node[state, right=of q2] (q3) {3};
    \node[state, right=of q3] (q4) {4};
    \node[state, accepting, right=of q4] (q5) {5};
    
    \path
      (q0) edge[loop, out=120, in=60, looseness=5] node[lab] {$\blk$:$\eps$} (q0)
      (q0) edge node[lab] {A:A} (q1)
    
      (q1) edge[loop, out=120, in=60, looseness=5] node[lab] {A:$\eps$} (q1)
      (q1) edge node[lab] {$\blk$:$\eps$} (q2)
      (q1) edge[bend right=25] node[lab, below] {$\del$:$\del$} (q3)
    
      (q2) edge[loop, out=120, in=60, looseness=5] node[lab] {$\blk$:$\eps$} (q2)
      (q2) edge node[lab] {$\del$:$\del$} (q3)
    
      (q3) edge[loop, out=120, in=60, looseness=5] node[lab] {$\del$:$\eps$} (q3)
      (q3) edge node[lab] {$\blk$:$\eps$} (q4)
      (q3) edge[bend right=25] node[lab, below] {B:B} (q5)
    
      (q4) edge[loop, out=120, in=60, looseness=5] node[lab] {$\blk$:$\eps$} (q4)
      (q4) edge node[lab] {B:B} (q5)
    
      (q5) edge[loop, out=120, in=60, looseness=5] node[lab] {B:$\eps$} (q5)
    ;
    
        \end{tikzpicture}
        \caption{Diacritic Decoding topology for the pattern A.B}
        \label{fig:ctc_modified}
    \end{subfigure}
    
    \caption{(a) Typical ASR CTC topology, where C represents the set of Arabic alphabetic characters, vs. (b) our proposed modified topology for constrained diacritic decoding for the pattern `A.B'. $\delta$ represents the wildcard characters (diacritics + special tokens).}
    \label{fig:decoding_graph}
\end{figure*}

\section{Background \& Related Work}
\label{sec:related_work}

\subsection{Arabic Diacritic Restoration}

Arabic diacritization has been studied extensively, but primarily as a text-based diacritic restoration problem, motivated by the fact that short vowels and other phonological marks are commonly omitted in everyday Arabic writing despite their importance for disambiguation \cite{habash_arabic_2007,belinkov_arabic_2015}. It is framed as a sequence labeling task and trained using different types of classical and neural architectures to predict diacritics corresponding to each input alphabetic character. 
A recent line of work investigates diacritics in the context of speech recognition, either by training ASR systems to emit diacritized text or by analyzing diacritic recognition behavior under different supervision conditions \cite{aldarmaki_diacritic_2023,alaqel_improving_2025}. These studies report that diacritic recovery can be improved when ASR is fine-tuned with manually diacritized transcripts and propose evaluation views that separate diacritic recognition quality from overall recognition accuracy \cite{aldarmaki_diacritic_2023}. Because fully diacritized Arabic speech data are scarce, prior works explored leveraging limited diacritized speech supervision in addition to text-based augmentation strategies, or using strong pretrained ASR models to bootstrap diacritized transcripts for subsequent multi-modal restoration models \cite{shatnawi-etal-2024-data,ghannam2025abjad}. In \cite{shatnawi_automatic_2024}, speech is utilized indirectly by augmenting the undiacritized text input with diacritized ASR hypotheses. These works demonstrate that incorporating speech as a supporting input modality besides text input can significantly improve diacritic restoration performance.

\subsection{Constrained Decoding and Lattice-based Inference}
Constrained decoding is a well-established idea in speech and language processing, with weighted finite-state transducers (WFSTs) providing a principled framework to encode pronunciation lexicons, grammars, and decoding constraints efficiently \cite{mohri_weighted_2002}. In neural generation, lexically constrained decoding methods (e.g., grid beam search) restrict outputs to satisfy required constraints without changing model parameters \cite{hokamp_lexically_2017}.  
Recent works extend CTC to incorporate partial or flexible supervision by expanding the set of admissible alignments \cite{starctc, gao_bypass_2023, gao_otc_2023}. For instance, \cite{nakagome_wctc-biasing_2025} applies wildcard CTC during decoding for \emph{biasing} and rare-word handling, using a wildcard-enabled scoring/search procedure to promote desired terms without retraining the base model. 
Mask-CTC \cite{higuchi_mask_2020} initializes a hypothesis using greedy CTC and then iteratively refines the sequence by masking low-confidence tokens and predicting them conditioned on the remaining context, combining CTC alignment with a mask-prediction objective. Our method is complementary to these: instead of refining arbitrary tokens, we impose a \emph{hard} transcript skeleton (the undiacritized characters) and restrict inference to diacritic insertions at permitted positions.

\begin{table}[tb]
    \centering
    \footnotesize
    \caption{Description of datasets used in our experiments. \textbf{CA} refers to Classical Arabic, \textbf{MSA} refers to Modern Standard Arabic (MSA), and (*) refers to text-only data. }
    \label{tab:datasets}
    \addtolength{\tabcolsep}{-0.4em}
\begin{tabular}{llcccc}
\multirow{2}{*}{\textbf{Variety}} & \multirow{2}{*}{\textbf{Dataset}} & \textbf{Train} & \textbf{Test} & \multirow{1}{*}{\textbf{Baselines}} & \multirow{1}{*}{\textbf{Ours}} \\
 & & \textbf{Size} & \textbf{Size}& §(\ref{sec:baseline})& §(\ref{sec:scenario_1})\\
\midrule
CA+MSA & Tashkeela* \cite{tashkeela}& - & -   & \checkmark \\
\midrule
CA & ClarTTS \cite{kulkarni2023clartts} & 12h & 0.3h  & \checkmark & \checkmark \\
\midrule
\multirow{1}{*}{MSA} & ArVoice\textsubscript{1,3} \cite{toyin25_interspeech} & 6h & 0.9h   & \checkmark & \checkmark\\

\midrule
\textbf{Total} & & 18h & 1.2h &  18h & 18h\\
\bottomrule
\end{tabular}
\end{table}

\section{Proposed Model}
\label{sec:methodology}
We address \emph{speech-to-text diacritization} in a setting where we are given the acoustic input utterance $x$ and the undiacritized character sequence $u = c_1 \ldots c_N$ as input. Our goal is to recover the missing diacritics using the speech signal while \emph{keeping the base characters fixed}, resulting in the diacritized character sequence $y = c_1 d_1 \ldots c_N d_N$. \\

\noindent\textbf{Label design.}
Each base character has at most one diacritic; multi-diacritic combinations are merged into a single-character token to preserve a 1-to-1 character-level representation. This is equivalent to most text-based diacritic restoration methods, formulated as a character-level sequence labeling task. In cases where the ground-truth labeling contains no diacritics, it is a common practice to use a special `no diacritic' token. In our method, we use the CTC blank token to denote the absence of diacritics.

\subsection{Acoustic Model and CTC Training}
\label{sec:ctc}

For training, we use available speech data with diacritized transcripts. We train a CTC-based ASR model to predict the target sequence of characters, which includes both alphabetical letters and their respective diacritics. Given acoustic feature sequence $x$, the neural encoder produces frame-level posteriors
\begin{equation}
p_t(k \mid x), \quad t = 1 \ldots \mathcal{T}, k \in \mathcal{V}\cup\{\epsilon\},
\end{equation}
where $\mathcal{V}$ is the output vocabulary and $\epsilon$ is the CTC blank symbol.
Under CTC, the probability of a label sequence $y$ is

\begin{equation}
P(y \mid x) = \sum_{\pi \in \mathcal{B}(y)}  P(\pi \mid x),
\end{equation}
where $\pi$ is a frame-level alignment path and $\mathcal{B}(\cdot)$ is the set of all valid paths $\pi$
 that map to $y$. 

Training minimizes the negative log-likelihood:
\begin{equation}
\mathcal{L}_{\mathrm{CTC}} = -\log P(y^\star \mid x),
\end{equation}
with $y^\star$ the reference diacritized transcript.\\

\subsection{Inference with Diacritization Constraint Lattice}
\label{sec:lattice}

In typical ASR inference, CTC decoding finds the most likely label sequence, and the label vocabulary in each time step includes all output characters, implemented as a Weighted Finite-State Transducer (WFST). The default CTC topology is presented in Figure \ref{fig:ctc_default}. In our application, we only need to predict the diacritic sequence while keeping the reference alphabetic sequence unchanged. To implement this contraint, we construct a lattice  $G_{\mathrm{char}}(u)$, which is a linear-chain WFST whose state track progress through the undiacritized transcript $u$, with additional wildcard states inserted after each Arabic letter $u_i$ for diacritic prediction. For instance, if the undiacritized transcript is $c_1c_2c_3$, the resulting pattern $p$ with wildcard (`.') characters will be `$c_1.c_2.c_3.$' .

Let $W\subseteq \mathcal{V}$ be the set of wildcard characters, which in our model are the composite Arabic diacritic labels in addition to the CTC blank symbol; the lattice is constructed as follows: for each character in $p$, 

\begin{itemize}
    \item If $p_j$ is a wildcard, then add parallel arcs for each wildcard characters in $W$
    \begin{equation}
    s_{j-1} \xrightarrow{\delta} s_{j},\quad \forall \delta \in W.
    \end{equation}

    \item Else, emit the character directly

    \begin{equation}
    s_{j-1} \xrightarrow{c_i} s_j.
    \end{equation}
\end{itemize}

This lattice compactly represents all diacritized strings compatible with $u$ and can be implemented as a WFST-style acceptor. The resulting decoding graph can be seen at \autoref{fig:ctc_modified}. 

Operationally, this can be realized by composing a CTC decoding graph with $G_{\mathrm{char}}(u)$, or by restricting beam expansions to only those symbols permitted by outgoing arcs in the lattice state corresponding to the current character position. Intuitively, this is a \textbf{partial forced-alignment} formulation: the decoding graph forces the base letters to appear in the output in the same order as in $u$, and the model only predicts which diacritic (if any) accompanies each letter.

\section{Experimental Details}
\label{sec:experiment}

\subsection{Datasets}
\autoref{tab:datasets} lists the datasets used to train and test the baseline and proposed models. We use two datasets covering different Arabic varieties to test the models' generalization capabilities:\\

\begin{enumerate}
    \item \textbf{ClArTTS} \cite{kulkarni2023clartts} provides carefully read single-speaker speech in Classical Arabic based on a classical religious book, with transcripts manually annotated with full diacritics. 
    \item \textbf{ArVoice} \cite{toyin25_interspeech} contains multi-speaker  read speech data from the news domain in Modern Standard Arabic; we utilize the fully diacritized portions (parts 1 and 3).
 
\end{enumerate}

\subsection{Data Preprocessing} 
We apply a lightweight normalization pipeline to standardize the text and reduce inconsistencies. Crucially, we normalize the character strings into the \textit{Normalization Form Canonical Composition (NFC)} form, as defined by the Unicode Standard, to unify the representation of the characters with multiple encodings (e.g., ``Alef with Hamza'' as a single character ``\textit{\textbackslash U0623}'' or multiple characters ``\textit{\textbackslash U0627\textbackslash U0654}''). This also standardizes the order of consecutive diacritics to a canonical order. We additionally collapse multiple consecutive whitespace characters into a single space.

\begin{table}
\centering
\footnotesize
\setlength{\tabcolsep}{3.1pt}
\renewcommand{\arraystretch}{1.08}
\caption{Performance of the ASR model from \cite{grosman2021xlsr53-large-arabic} before and after fine-tuning on the combined training splits of ClArTTs and ArVoice. WER/CER are calculated using diacritized texts (\textbf{Diac}) and undiacritized texts (\textbf{Undiac}) to isolate diacritic errors from base character errors. Reference Diacritic Coverage (DCov): ArVoice=73.19, ClarTTS=79.13.}
\label{tab:ASR_results}
\begin{tabular}{llccccc}
\toprule
\multirow{2}{*}{\textbf{Model}} & \multirow{2}{*}{\textbf{Test Set}} & \textbf{DCov} &
\multicolumn{2}{c}{\textbf{Diac}} & \multicolumn{2}{c}{\textbf{Undiac}} \\
 &  & (\%) & \textbf{WER} & \textbf{CER} & \textbf{WER} & \textbf{CER} \\
\midrule
\multirow{2}{*}{w2v - zero shot}& ArVoice & 50.55 & 88.58 & 24.21 & 44.02 & 10.40 \\
& ClArTTS & 70.19 & 58.91 & 15.27 & 24.08 & 7.02 \\
\midrule
\multirow{2}{*}{w2v - fine-tuned} & ArVoice & \textbf{75.25} & \textbf{40.82} & \textbf{6.51} & \textbf{19.94}  & \textbf{3.61} \\ & ClArTTS & \textbf{79.12} & \textbf{17.17} & \textbf{3.17} & \textbf{10.03}  & \textbf{2.46} \\

\bottomrule
\end{tabular}
\end{table}

\subsection{Baselines}
\label{sec:baseline}

For baselines, we use the two models described in \cite{shatnawi_automatic_2024}:

~\\
\noindent\textbf{Text-only Model:} This model is composed of a two-layer Transformer text encoder to process the input character sequence, and a linear classification head to predict a diacritic label for each character position.
%
This model is first trained on the fully diacritized text transcripts of Tashkeela, which contain a mixture of CA and MSA samples (but mostly CA), and is then further fine-tuned on the target training split used in our experiments. This baseline represents the standard text-based diacritic restoration approach, where no speech data are used to infer the missing diacritics.

\begin{figure*}[t]
\centering

\begin{minipage}[t]{0.56\textwidth}
\vspace{0pt}
\centering
\captionof{table}{Cross-dataset results in terms of Word Error Rate (WER) and Diacritic Error Rate (DER) compared to baselines from \cite{shatnawi_automatic_2024}. DER includes ``no diacritic'' and word endings.}
\label{table:main_results}
\resizebox{\linewidth}{!}{%
\begin{tabular}{|l|c|cc|cc|}
\cline{2-6}
\multicolumn{1}{c|}{} &
\textbf{Test Set $\rightarrow$} &
\multicolumn{2}{c|}{ClArTTS} &
\multicolumn{2}{c|}{ArVoice} \\
\hline
\textbf{Model} &
\textbf{Train Set $\downarrow$} &
\textbf{WER} & \textbf{DER} &
\textbf{WER} & \textbf{DER} \\
\hline

Text Baseline & \multirow{3}{*}{ClArTTS} & 33.94& 11.7 & 80.05 & 39.97 \\
\cline{1-1}\cline{3-6}
Text +ASR Baseline & & 12.33& 3.54 &56.20 & 19.21 \\
\cline{1-1}\cline{3-6}
\textbf{Ours} & &\textbf{11.21} & \textbf{3.53} &\textbf{39.89} & \textbf{12.04}\\
\hline

Text Baseline & \multirow{3}{*}{ArVoice} &72.43 & 30.03 &62.19 & 21.05 \\
\cline{1-1}\cline{3-6}
Text +ASR Baseline & &99.94 & 76.61 & 40.17 & 11.9 \\
\cline{1-1}\cline{3-6}
\textbf{Ours} & &\textbf{34.94} & \textbf{11.86} & \textbf{27.87}& \textbf{7.73}\\
\hline

Text Baseline & \multirow{3}{*}{\shortstack{ClArTTS\\+ ArVoice}} & 31.82& 11.87 & 55.86 & 16.43  \\
\cline{1-1}\cline{3-6}
Text +ASR Baseline & & 29.63 & 9.05 & 34.47& 9.93 \\
\cline{1-1}\cline{3-6}
\textbf{Ours} & &\textbf{13.05} & \textbf{3.80} & \textbf{30.36} & \textbf{8.69} \\
\hline

\end{tabular}%
}
\end{minipage}
\hfill
\begin{minipage}[t]{0.40\textwidth}
\vspace{0pt}
\centering
\includegraphics[width=\linewidth]{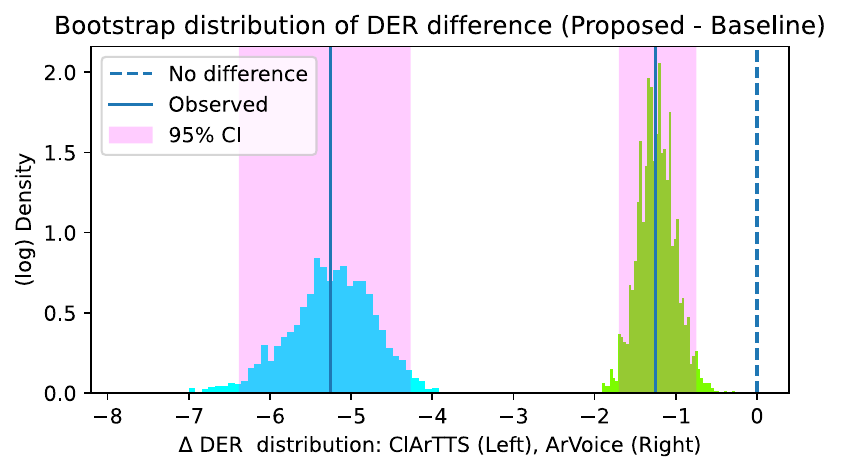}
\captionof{figure}{The bootstrap distribution of the differences in DER between our method (proposed) and text+ASR (baseline), trained on the combined ClArTTS and ArVoice sets, with 95\% confidence intervals.}
\label{fig:bootstrapping}
\end{minipage}

\end{figure*}

~\\
\noindent\textbf{Text+ASR Model:} This model augments the text-only transformer with information derived from speech. A pre-trained ASR model first produces a diacritized hypothesis from the input speech utterance. The undiacritized transcript and the diacritized ASR hypothesis are then encoded by two separate but identically configured transformer encoders. The two representations are then fused via cross-attention at the final layer: the representations of the undiacritized transcript are used for the \emph{query} vectors, and the representations of the ASR hypothesis are used for the \emph{key} and \emph{value} vectors. This yields fused representations aligned to the length of the undiacritized transcript. A linear classification head is then applied to the fused representations to predict a diacritic label for each character position.
The model's text encoder is first initialized from the Tashkeela-based \textbf{Text-only~Model} described above before fine-tuning the full multi-modal diacritic restoration system.  

\subsection{Experimental Settings}
\label{sec:scenario_1}

\noindent\textbf{Evaluation Metrics:}
We report word-error-rate (WER) and character-error-rate (CER) metrics for both ASR and diacritic restoration models. We also report the \textbf{Diacritic Coverage Rate (DCov)}, which measures the pervasiveness of diacritics in a system's output. It is defined as the percentage of base Arabic characters that are diacritized out of the total number of base Arabic characters in a sentence.

To assess diacritic restoration performance, we report \textbf{Diacritic Error Rate (DER)},
measuring the fraction of characters for which the predicted diacritic differs from the reference, under the \emph{strict} assumption that the base-character sequences in the reference and hypothesis match (i.e., diacritics are compared at the same character positions). 
we report the total DER including \textit{no diacritic} and including \textit{word ending} diacritics\footnote{In Arabic, word ending diacritics often denote grammatical rather than morphological information. We do not make a distinction here.}.\\

\noindent\textbf{ASR Model:} We use a Wav2vec2-XLSR \cite{grosman2021xlsr53-large-arabic} model originally trained on Arabic speech from the Common Voice and Arabic Speech Corpus. We fine-tune it using the relevant training corpus for each experiment. We report the baseline ASR performance in \autoref{tab:ASR_results} using standard CTC decoding, before and after fine-tuning it on the combined training set. Fine-tuning improves both diacritics coverage and error rates.

~\\
\noindent\textbf{Experiment's Settings and Objectives:} The ASR model used in the baseline and ours are always identical in our experiments, providing a fair basis for comparison. Note also that the baseline has the advantage of incorporating additional data from the text domain (i.e., the Tashkeela corpus), whereas our model cannot utilize such data. Our model, however, is more compact and streamlined as it directly utilizes the ASR model without the need for additional text encoders. 

~\\\noindent\textbf{Optimization:} We fine-tune all models using AdamW optimizer with a linear learning-rate schedule (warmup followed by decay), using a peak learning rate of $3e^{-4}$ and 1,500 warmup steps. Training runs for 100 epochs with a batch size of 64 on an NVIDIA A100 GPU (80GB VRAM).

\section{Results and Discussion}

Table \ref{table:main_results} shows diacritic restoration results of our proposed method compared to the baselines under matched, cross-dataset, and combined-training settings to thoroughly evaluate generalization performance. First, comparing the \textbf{Text-only} and \textbf{Text+ASR} baselines, our results conform with previous findings from \cite{shatnawi_automatic_2024} that the speech modality significantly improves diacritic restoration performance. In the matched ClArTTS setting, the Text+ASR baseline and our method achieve nearly identical DER. 
However, our method shows stronger robustness under dataset mismatch. When trained on ClArTTS and evaluated on ArVoice or vice versa, the Text+ASR baseline degrades sharply, whereas our method shows better robustness and generalization. This suggests that relying only on the final ASR hypothesis can lead to poor cross-dataset generalization, while using richer information from the CTC decoding graph yields more consistent performance.

The combined-training setting further supports this trend. When trained on both ClArTTS and ArVoice, our method achieves the best performance on both test sets. To verify the statistical validity of this improvement, we applied non-parametric bootstrap resampling to report confidence intervals. Specifically, we generated 2,000 bootstrap samples with replacement, and we computed the performance difference between our proposed method and the Text+ASR baseline in the combined-training setting. The empirical distribution of these differences was used to estimate the 95\% confidence intervals, which are plotted in \autoref{fig:bootstrapping}. In this figure, negative values indicate that the proposed method achieves lower DER compared to the baseline for a specific boostrap sample. Note that for both datasets, the confidence intervals lie entirely below zero, indicating that the improvement in DER is significant.

\section{Conclusion}

We described a constrained CTC decoding method for Arabic diacritic restoration and demonstrated its effectiveness against comparable baselines. The approach results in better performance and better generalization across two datasets of Modern Standard and Classical Arabic, while being more streamlined and efficient as it does not rely on text fusion or multi-modal training.  
This approach enables us to use the ASR model directly as either a fully-fledged diacritized ASR model for transcribing unlabeled data, or as a constrained diacritic restoration model for speech data that are labeled with undiacritized reference texts. Considering that the majority of existing speech datasets for Arabic speech have undiacritized transcripts, the model can be used as a robust diacritic restoration method to complement speech data curation efforts for Arabic.

\section{Generative AI Use Disclosure}
We confirm that Generative AI tools were not used to produce any part of this manuscript.

\bibliographystyle{IEEEtran}
\bibliography{mybib}


\section{Appendix}

\appendix
\section{Extending to partially diacritized data}\label{sec:appendix-a}

\begin{table*}[thb]
\centering
\caption{Speaker source and diacritic coverage rate (DCov) in training and testing subsets for seen and unseen speakers. Duration is reported in hours. \textcolor{blue}{$\star$} denotes modified diacritic coverage for experimental setups 2 and 3. \textsuperscript{$\dagger$} denotes Synthetic voice.}
\label{tab:speaker-distribution}
\begin{tabular}{clcccccc}
\toprule
&\multirow{2}{*}{\textbf{Speaker ID}} &
\multirow{2}{*}{\textbf{Source}} & \multicolumn{2}{c}{\textbf{Dcov (Train)}}& \textbf{Dcov (Test)} & \multicolumn{2}{c}{\textbf{Duration (h)}}\\
&& & Setup-1& Setups-2/3 & Setups-1/2/3 & Train & Test\\
\midrule
\multirow{7}{*}{\textbf{Seen}}&female\_ad & ArVoice P1 & 73.11 & 73.11 & 72.93 & 1.10 & 0.13\\
&male\_asc & ArVoice P3 & 76.86 & 76.86 & 70.18 & 2.41 & 0.29\\
&male-Wavenet-B\textsuperscript{$\dagger$} & Arvoice P4 & 73.06 & 73.06 & - & 1.43 & 0.14 \\
&female-Wavenet-D\textsuperscript{$\dagger$} & Arvoice P4  & 74.81 & \textcolor{blue}{$\star$} 36.31 & - & 1.30 & 0.15 \\
&male\_aa & ArVoice P1 & 74.02 & \textcolor{blue}{$\star$} 37.95 & 74.54 & 1.02 & 0.14\\
&male\_ae & ArVoice P2 & - & 0 & - & 0.91 & - \\
&female\_af & ArVoice P2 & - & 0 & - & 0.82 & - \\
\midrule
\multirow{2}{*}{\textbf{Unseen}}&male\_ac & ArVoice P1 & - & - & 74.30 & - & 0.14\\
&female\_ab & ArVoice P1 & - & - & 73.72 & - & 0.15\\
\bottomrule
\end{tabular}
\end{table*}

In this appendix, we examine how the diacritic coverage of the ASR model can be affected when the training data contains a mixture of fully diacritized, partially diacritized, and undiacritized sources. While the main experiments assume fully diacritized supervision, this assumption does not always hold in realistic Arabic ASR settings, where much of the available data is only partially diacritized or lacks diacritics entirely. Incorporating such data is important for improving ASR robustness and generalization across speakers and domains; however, it can also bias the model toward producing undiacritized or sparsely diacritized outputs. Since our diacritic restoration approach relies on the ASR model, the learned diacritic distribution directly affects the coverage and reliability of the restored diacritics. We therefore study whether diacritic coverage can be improved under mixed-coverage training conditions, without degrading the diacritic prediction quality.

\begin{figure}[thb]
    \includegraphics[]{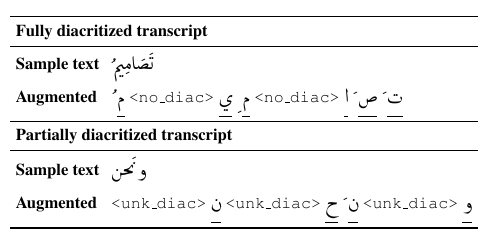}
    \caption{An example illustrating how the additional \texttt{<no\_diac>} and \texttt{<unk\_diac>} tokens were added. In fully diacritized transcripts, \texttt{<no\_diac>} is inserted where the transcript does not provide diacritics, while in partially diacritized transcripts, \texttt{<unk\_diac>} is inserted when the diacritic label is not available.}
    \label{fig:diacritic-token-augmentation-example}
\end{figure}

We hypothesize that diacritic coverage would be degraded if the training data include a large portion of characters with `no diacritic' label. Fully diacritized datasets also include cases where the gold label is to have no diacritics. We propose to treat these differently from cases where the true diacritic is unknown. To this end, we introduce two distinct wildcard tokens to distinguish the two cases: \texttt{<no\_diac>} to denote cases where the true label is to have no diacritic at this position, which we use in the case of fully diacritized data; \texttt{<unk\_diac>} to denote cases where the true diacritic at this position is unknown, which we use for partially diacritized and undiacritized data. See \autoref{fig:diacritic-token-augmentation-example} for an example. During inference, we exclude \texttt{<unk\_diac>} from the set of wildcard tokens to enforce the prediction of diacritics that match the distribution of fully diacritized subsets of the training corpus.

\subsection{Train/Test Configuration}

For this experiment, we construct the train and test subsets from the four parts of the ArVoice dataset to represent a mixture of speaker conditions and diacritic coverage levels. Speakers \texttt{male\_aa}, \texttt{male\_asc} and \texttt{female\_ad} are included in both training and testing to evaluate seen speaker performance, while \texttt{male\_ac} and \texttt{female\_ab} are held out from training and used only for testing to evaluate generalization to unseen speakers. To preserve coverage of the corresponding utterance content in training while maintaining these two speakers as unseen test speakers, we replace \texttt{male\_ac} and \texttt{female\_ab} in the training subset with their parallel synthetic voices from ArVoice Part 4, \texttt{male-Wavenet-B} and \texttt{female-Wavenet-D}, respectively. This results in a training set that includes both natural and synthetic speaker sources while keeping \texttt{male\_ac} and \texttt{female\_ab} unseen during training.

To simulate training data with mixed diacritic availability, we modify only the training transcripts while keeping the test set fixed. The training set contains speakers with naturally high diacritic coverage, speakers with no diacritic supervision from the undiacritized portion of ArVoice Part 2, and speakers whose coverage is artificially reduced to represent partially diacritized sources. Specifically, we randomly remove diacritics from the training transcripts of \texttt{male\_aa} and \texttt{female-Wavenet-D}, reducing their diacritic coverage from 74.0\% to 37.9\% and from 74.8\% to 36.3\%, respectively. \autoref{tab:speaker-distribution} shows the distribution of train and test subsets by speaker. 

\subsection{Experimental Setup}

We compare the following three setups: 
~\\

\noindent\textbf{Setup-1: Upperbound (full)} We train on fully diacritized data only\footnote{We do not include the undiacritized ArVoice Part 2 in this setup. } and test with constrained decoding (no additional special characters). This is similar to the original setup described in this paper and serves as an upper-bound reference for diacritic coverage when training data distribution contains only fully diacritized sources.
\\

\noindent\textbf{Setup-2: Naive (partial)} We train the model using the mixed-coverage subsets described in \autoref{tab:speaker-distribution} and test constrained decoding with no additional special characters. This setup demonstrates the effect of training data distribution on the resulting diacritic coverage rate. 
\\

\begin{table*}[thb]
\centering
\caption{Diacritic Coverage Rate (Dcov) and Diacritic Error Rate (DER) for seen and unseen speakers from the ArVoice test set across the three setups, including and excluding the `no diacritic' token (n/d) in prediction.}
\label{tab:results}
\begin{tabular}{|c|l|c|cc|}
\cline{2-5}
 \multicolumn{1}{c|}{}& \multirow{2}{*}{\textbf{Setup}}& \multirow{1}{*}{\textbf{Dcov (\%)}}
 & \multicolumn{2}{c|}{\textbf{DER (\%)}} \\
 \cline{4-5}
  \multicolumn{1}{c|}{}&
  & \textbf{(Predictions)} & \textbf{inc. n/d} & \textbf{exc. n/d} \\
\hline
\multirow{3}{*}{\rotatebox{90}{\textbf{seen}}}
 & Setup-1: Upperbound (full) & 75.92 & 8.77 & 9.79 \\
 & Setup-2: Naive (partial) & 59.75 & 22.57 & 10.59 \\
 & Setup-3: Special-characters (partial) & 71.50 & 12.18 & 10.02 \\
\hline
\multirow{3}{*}{\rotatebox{90}{\textbf{unseen}}}
 & Setup-1: Upperbound (full) & 75.52 & 9.22 & 8.80 \\
 & Setup-2: Naive (partial) & 53.13 & 30.44 & 9.35 \\
 & Setup-3: Special-characters (partial) & 73.17 & 10.99 & 8.98 \\
\hline
\end{tabular}
\end{table*}

\begin{figure*}[thb]
\centering
\includegraphics[width=0.95\textwidth
    ]{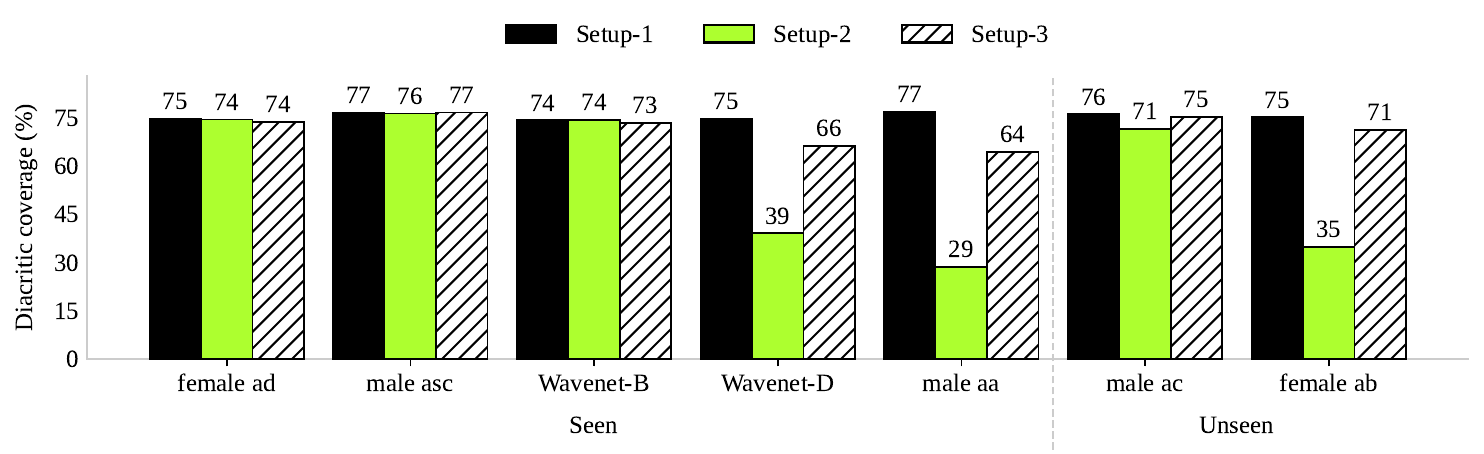}
\captionof{figure}{Diacritic coverage rate per speaker (seen and unseen) in the three experimental setups, on the test set. \textbf{Note:} speakers \texttt{male\_ae} and \texttt{female\_af} are not included in the evaluation, as their transcripts are undiacritized in the train and test splits.}\label{fig:coverage}
\end{figure*}

\noindent\textbf{Setup-3: Special-characters (partial)} Similar to Setup-2, but trained with the \texttt{$<$no\_diac$>$} and \texttt{$<$unk\_diac$>$} special characters. At inference, we exclude \texttt{$<$unk\_diac$>$} from the set of wilcard tokens. This demonstrates the  effect of the proposed method for controlling diacritic coverage rate under mixed coverage conditions. \\

For all setups, we use the same settings described in \S\ref{sec:scenario_1} and the same ASR model; we only modify the training data diacritic distribution and inference setup.

\subsection{Results}
\vspace{5.8pt}
\noindent\textbf{Effect of partially diacritized training data on diacritic coverage.} 
In setups 1 and 2, the distribution of the diacritic coverage for seen speakers on the test set (\autoref{fig:coverage}) follows their distribution in the training set (\autoref{tab:speaker-distribution}). The speakers with partial diacritics (\texttt{male\_aa}, \texttt{female-Wavenet-D}) in setup-2 resulted in similar partial coverage in the test set.

For the unseen speakers, the diacritic coverage in setup 2 for \texttt{female\_ab} is reduced to 35\% compared to 75\% in setup-1. Interestingly, for the unseen male speaker, \texttt{male\_ac}, the coverage matches the baseline. Multiple confounds could be the cause of this interesting behaviour, one of which is the higher representation of fully diacritized male voices in the training set compared to female voices, particularly the \texttt{male\_asc} subset, which is larger than the rest \cite{toyin25_interspeech}.

Overall, these results demonstrate that higher proportions of undiacritized or partially diacritized data in training would likely reduce diacritic coverage at test time for unseen target speakers/domains, when naively mixed with fully diacritized data. This limits the applicability of the model, in the naive setup, for full coverage diacritic restoration. 
\\

\noindent\textbf{Maximizing diacritic coverage using the special tokens.} Compared to setup-2, our results in setup-3 illustrate that using the special tokens \{\texttt{<no\_diac>}, \texttt{<unk\_diac>}\} allows us to include partially diacritized and undiacritized data (crucial for domain coverage) when training the ASR model, without hurting the diacritic coverage rate at inference. The experiment shows consistent train-test diacritic coverage rates for seen speakers with lower coverage (e.g., \texttt{male\_aa}) as well as unseen speakers (e.g., \texttt{female\_ab}). 
\\

\noindent\textbf{Impact of increased coverage on diacritic error rates.} \autoref{tab:results} shows the DER across setups. We show the results, including and excluding characters with no diacritics in the \textit{predicted} transcripts. This enables us to measure the precision of diacritic restoration even when the model only produces partial diacritics (as in setup 2) without penalizing the model for omitting diacritics. It also enables us to measure whether the forced additional diacritic coverage in setup-3 is happening at the cost of accuracy, which would be counterproductive. 

Note that setup-1 is trained with the fully diacritized data, whereas setups 2 and 3 are trained with partially diacritized data, so diacritic error rate increase in the latter two is expected. More importantly, DER in setup-3 is not higher than that of setup-2, which indicates that the additional diacritics introduced by excluding the unknown token match the quality of the diacritics predicted by default, so the quality of the output is not compromised while improving coverage.

These results overall illustrate the effectiveness of the proposed method for diacritic restoration even in cases where the training set includes undiacritized and/or partially diacritized sources, making it more practical and generalizable.

\end{document}